# Data Excellence for AI: Why Should You Care

Lora Aroyo*[1], Matthew Lease[2,3], Praveen Paritosh[1], Mike Schaekermann[2]
[1]Google Research, [2]Amazon, [3]University of Texas at Austin

## Data is the fuel and the compass for AI, yet research on data receives little attention

The efficacy of machine learning (ML) models depends on both algorithms and data. Training data defines what we want our models to learn, and testing data provides the means by which their empirical progress is measured. Benchmark datasets define the entire world within which models exist and operate, yet research continues to focus on critiquing and improving the algorithmic aspect of the models rather than critiquing and improving the data with which our models operate. If "data is the new oil," we are still missing work on the refineries by which the data itself could be optimized for more effective use.

**Data is potentially the most under-valued and de-glamorised aspect of today's AI ecosystem**. Data issues are often perceived and characterized as mundane "pre-processing" that has to be done before getting to the real (modeling) work. ML practitioners often view data wrangling as tedious and time-consuming. In contrast, Sambasivan et al. (2021) provide examples of how data quality is crucial to ensure that AI systems can accurately represent and predict the phenomenon it is claiming to measure. Amershi et al, (2019) identify three aspects of the AI domain that make it fundamentally different from prior software application domains. One of the challenges identified focuses on processes for data discovery, management, and versioning, which are much more complex and difficult than in traditional software engineering.

**Real-world datasets are often 'dirty'**, with various data quality problems and present the risk of "garbage in = garbage out" in terms of the downstream AI systems we train and test on such data. This has inspired a steadily growing body of work on understanding and improving data quality. It also highlights the importance of rigorously managing data quality using mechanisms specific to data validation, instead of relying on model performance as a proxy for data quality. Just as we rigorously test our code for software defects before deployment, we need to test for data defects with the same degree of rigor, so that we could detect or prevent weaknesses in ML models caused by underlying issues in data quality.

**Benchmark datasets are often missing much of the natural ambiguity of the real world,** as data instances with annotator disagreement are often aggregated to eliminate disagreement (obscuring uncertainty), or filtered out of datasets entirely. Successful benchmark models fail to generalize real data, and inflated benchmark results mislead our assessment of state-of-the-art capabilities. Thus, ML models become prone to develop "weak spots", i.e., classes of examples that are difficult or impossible for a model to accurately evaluate, because that class of examples is missing from the evaluation set (Aroyo et al, 2021).

**Measuring data quality is challenging, nebulous, and often circularly defined with "ground truth"** on which models are tested. When dataset quality is considered, the ways in which it is measured in practice is often poorly understood and sometimes simply wrong. Challenges identified include fairness and bias issues in labeled datasets, quality issues in datasets, limitations of benchmarks, reproducibility concerns in machine learning research, and lack of documentation and replication of data.

Measurement of AI success today is often metrics-driven, with emphasis on rigorous model measurement and A/B testing. However, measuring the goodness of the fit of the model to the dataset ignores any consideration of how well the dataset fits the real world problem. Goodness-of-fit metrics, such as F1, Accuracy, AUC, do not tell us much about data *fidelity* (i.e., how well the dataset represents reality) and *validity* (how well the data explains things related to the phenomena captured by the data). No standardised metrics exist today for characterising the goodness-of-data.

## How can lessons from excellence in Software Engineering help Data Excellence?

Decades of catastrophic failures within high-stakes software projects have helped to establish the crucial importance of upfront investments in *software engineering excellence*. It was through careful post-hoc analysis of these kinds of disasters that software engineering has matured as a field and achieved a robust understanding of the costs and benefits: *processes* like systematic code reviews, *standards* like coding conventions and design patterns, *infrastructure* for debugging and experimentation, as well as *incentive structures* that prioritize careful quality control over hasty roll-outs. An analogous framework for *data excellence* does not yet exist, bearing the risk of similarly disastrous catastrophes to arise from the use of datasets of unknown or poor quality in AI systems. Consider some **key properties that can pave the way for data excellence in analogy with software engineering excellence**:

**Maintainability:** Maintaining data at scale has similar challenges as maintaining software at scale. We can follow the lessons learned from software engineering and apply them to data and its maintenance.

**Validity** tells us about how well the data helps us explain things related to the phenomena captured by the data, e.g., via correlation between the data and external measures. Education research tries to explore whether grades are valid by studying their correlations to external indicators of student outcome. For datasets to have operational validity we need to know whether they account for potential complexity, subjectivity, multi-valence or ambiguity of the intended construct; whether they can predict features of the represented construct; and whether the data collection procedure allows for generalizations within the intended phenomena.

**Reliability** captures internal aspects of data validity, such as: consistency, replicability, reproducibility of data. Irreproducible data allows us to draw whatever conclusions we want to draw, while giving us the facade of being data-driven, when it is dangerously hunch driven. We need reliable mechanisms to account for the human aspects of data collection.

**Fidelity:** Users of data often assume that the dataset accurately and comprehensively represents the phenomenon, which is almost never the case. For example, various types of sampling from larger corpora can impact the fidelity of a dataset, e.g. temporal splitting can introduce bias if not done right in cases such as signal processing or sequential learning; or user-based splitting not keeping data of users separated is a potential bias source (e.g. if data from the same user is in test and train sets).

# Report from the 1st Data Excellence Workshop @ HCOMP2020

Researchers in human computation (HCOMP) and ML-related fields have shown a longstanding interest in human-annotated data for model training and testing. A series of workshops (Meta-Eval2020 @AAAI, REAIS2019 @HCOMP, SAD2019 @TheWebConf, SAD2018 @HCOMP) raised awareness about the issues of data quality for ML evaluation and provided a venue for related scholarship. As human-annotated data represents the compass that the entire ML community relies on, data-focused research can potentially have a multiplicative effect on accelerating progress in ML broadly.

The 1st Data Excellence Workshop @HCOMP2020 provided a platform for an interdisciplinary group of researchers from industry and academia to discuss and inspire a framework for data excellence in AI systems. Over 100 participants from 60 international institutions participated. At the outset of this workshop, we proposed that **data is "excellent" when it *accurately* represents a phenomenon, and is: (a) collected, stored, and used responsibly; (b) maintainable over time; (c) reusable across applications; (d) exhibits empirical and explanatory power.** Leveraging lessons learned from decades of software engineering inspired an analogous framework for data excellence in AI systems with respect to:
- Identifying properties and defining metrics for data excellence
- Gathering case studies with examples of data excellence and data catastrophes
- Cataloging best practices and incentive structures for data excellence
- Discussing the cost-benefit trade-off for investments in data excellence
- Cataloging methodologies for reliability, validity, maintainability, fidelity of data

Seven Invited talks and three contributed papers covered research and practical experiences from a wide range of industry, academia and government organizations. They contributed case studies of both data catastrophes and excellence, including empirical and theoretical methodologies for reliability, validity, maintainability, and fidelity of data (See Fig1 and Fig2).

## Invited Talks

*Emily Dinan* discussed data challenges for neural dialogue models which may learn and mimic patterns of offensive or otherwise toxic behavior when trained on large unlabeled corpora of human interactions. *Aleksander Mądry* presented results in support of the notion that ML models pick up biases of the real-world world and our data pipelines. *Quang Duong* introduced the complex task of medical data labeling and highlighted several differences from traditional crowdsourced labeling such as: increased task complexity and duration, the requirement of medical expertise in human graders, additional personas like workforce managers who not only coordinate expert labelers and oversee data quality, but also contribute to the development of complex grading guidelines and training of graders. *Andrea Olgiati* discussed the evolution of best practices in software engineering and drew parallels to the process of dataset creation for AI applications, e.g. unit tests for dataset creation ought to verify both the syntax (are labels in the right format?) and semantics (do labels have the right meaning and magnitude?) of data labels. *Ian Soboroff* provided a historical overview of datasets and data evaluation methods produced by NIST's Text Retrieval Conference (TREC). In this context, Soboroff called out the need for "relentless introspection", i.e., building and scrutinizing a dataset within the community of use. *Peter Hallinan* defined data excellence as an optimal tradeoff between dataset quality, cost and speed, given fixed constraints (privacy and product requirements). He further suggested that dataset quality, cost and speed are system properties resulting from design levers that can be controlled, including portfolio-level and dataset-level control levers. *Ben Hutchinson* emphasized the need to view datasets as "engineered infrastructure." He introduced a rigorous framework for dataset development transparency which supports decision-making and accountability.

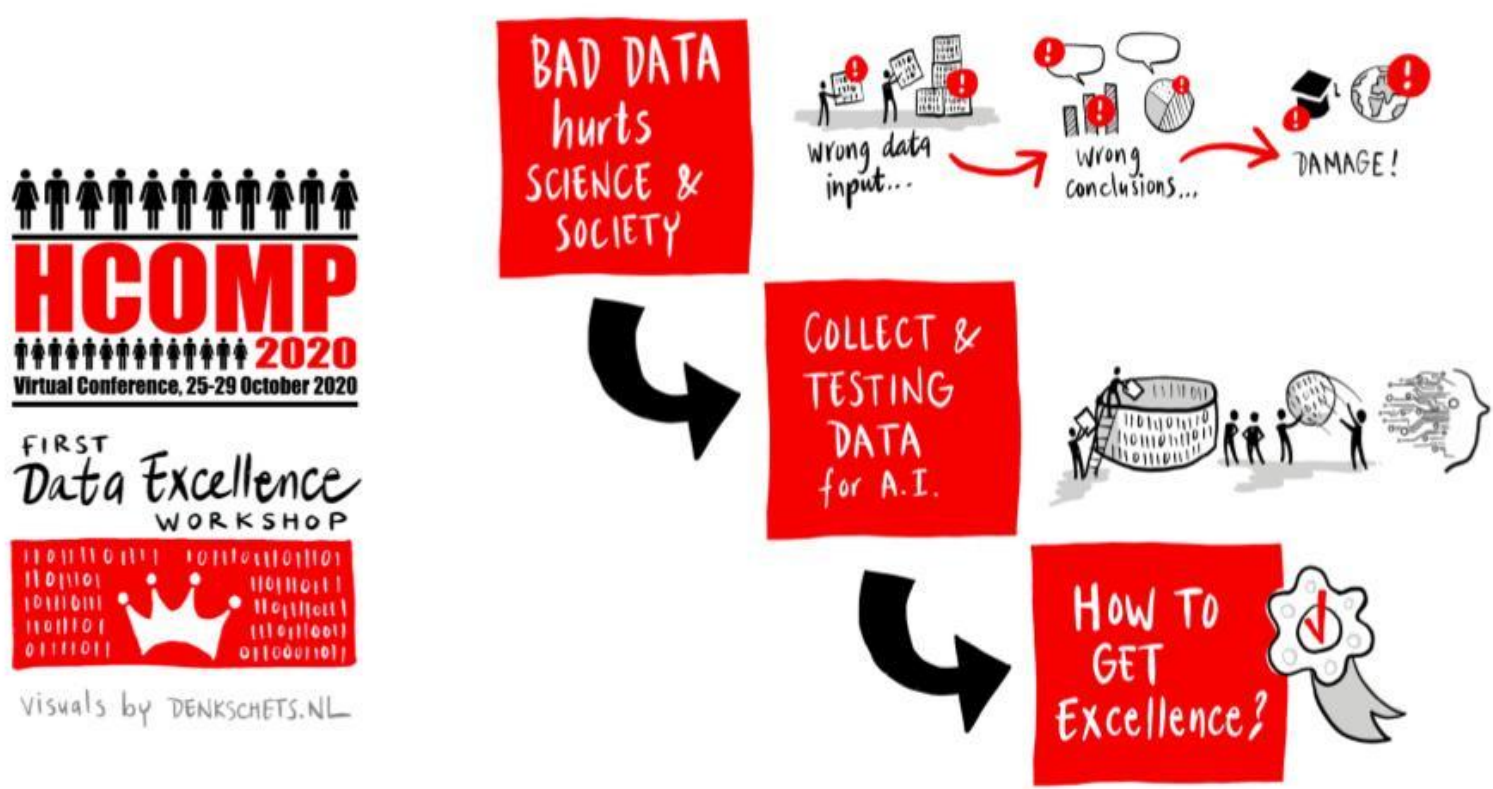

**Figure 1: Data Excellence Workshop: visual summary**

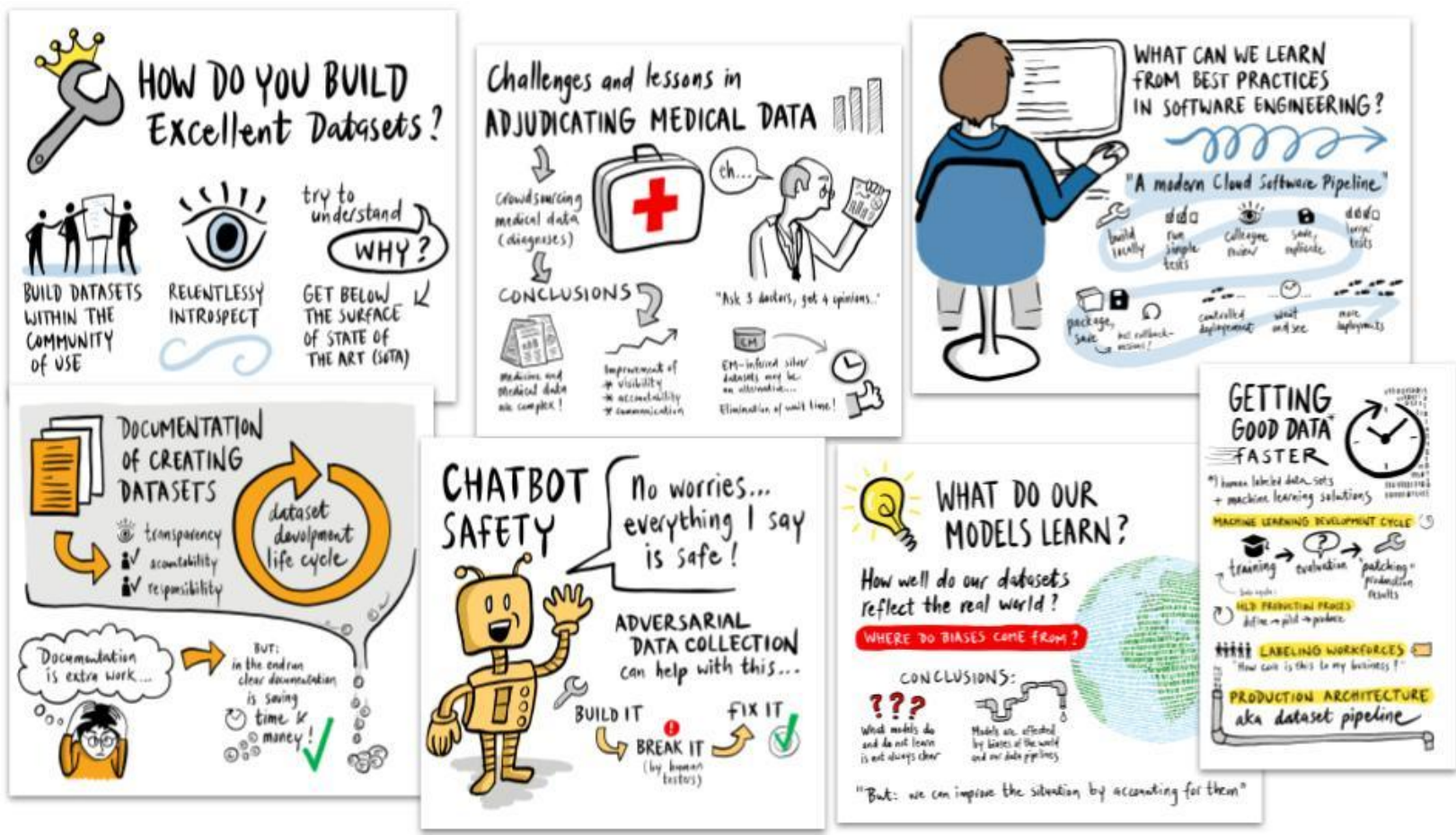

Figure 2: Data Excellence workshop: paper visual overviews

## Papers

*Han et al. 2020* define "annotation artifacts" as a type of dataset bias in which annotations capture workers' idiosyncrasies that are irrelevant to the task itself. Their work provided empirical results to address the questions of what factors affect the generation of annotation artifacts and how to reduce annotation artifacts by adjusting workflow design. *Christensen et al. 2020* advocated for new methods of producing labeled training data for machine learning that can discover and support diverse opinions and flexible problem solving to serve all users equitably. *Kapania et al. 2020* report on data practices in high-stakes AI, from interviews with 53 AI practitioners in India, East and West African countries, and the USA. Further, they report on the challenges faced by practitioners with two essential desiderata within high-stakes domains: data reliability and fidelity and discuss strategies for their improvement and to avoid data disasters—resulting in safer and robust systems for all.

## Discussion

**Data as Science vs. Data as Engineering.** Data intersects software engineering and scientific inquiry. Both interpretations of data as engineering and as science emerged as productive metaphors that can teach us about the central questions of data excellence, but possibly point into different directions: engineering is goal-focused, science is curiosity-focused; engineering is more strongly represented in industry, science more strongly in academia; engineering has less theory and more practice, and science is the other way. Discussions highlighted open questions for both perspectives:

Data as Science:
- How can we as a community address the reproducibility crisis in data collection for AI systems which has implications throughout the life cycle of AI systems including model design, development, evaluation and deployment.
- How can we make progress on measuring AI model performance on the underlying task we care about as opposed to the benchmark itself?

Data as Engineering:
- How can we best leverage synergies between human judgment and algorithmic approaches like low-shot and unsupervised learning in the process of annotating data?
- How can we formalize data requirements specification, so it is consistent, repeatable, and well-defined for others?
- Which stakeholders are most concerned with data fidelity?

**Best practices.** The workshop discussion also addressed questions of establishing, communicating and incentivizing best practices towards data excellence both for novices and experienced practitioners alike, and highlighted open questions:

Datasets as Living Artifacts:
- What infrastructure, tools and practices can facilitate maintaining datasets that are alive (non-static) and grow over time?
- For continuously evolving benchmark datasets that are collected with models-in-the-loop, how can we prevent the dataset from drifting away from some reasonable distribution for the original task?

Documentation:
- How much dataset documentation is enough, and how is dataset documentation justified when other priorities are competing for attention on a pressing deadline?
- What are best practices in dataset design documents for discussing the biases your dataset might have and how others might help address those biases?

## Conclusion

Optimizing the cost, size, and speed of collecting data has attracted significant attention in the *first-to-market* rush with data for AI. However, important aspects of maintainability, reliability, validity, and fidelity of datasets have often been neglected. We argue we have now reached an inflection point in the field of ML in which attention to neglected data quality is poised to significantly accelerate progress. We advocate for research defining and creating processes to achieve *data excellence* and showcase examples, case-studies, and methodologies *to enable a shift in our research culture to value excellence in data practices, to enable the next generation of breakthroughs in ML and AI*.

## References

**Workshop Accepted Papers**
- Data Desiderata: Reliability and Fidelity in High-stakes AI, presented by Shivani Kapania with co-authors Nithya Sambasivan, Kristen Olson, Hannah Highfill, Diana Akrong, Praveen Paritosh and Lora Aroyo.
- Machine Learning Training to Support Diversity of Opinion, presented by Johanne Christensen with co-author Benjamin Watson.
- Reducing Annotation Artifacts in Crowdsourcing Datasets for Natural Language Processing, presented by Donghoon Han with co-authors Juho Kim and Alice Oh.

**Additional References**
- Saleema Amershi, Andrew Begel, Christian Bird, Robert DeLine, Harald Gall, Ece Kamar, Nachiappan Nagappan, Besmira Nushi, and Thomas Zimmermann. 2019. Software engineering for machine learning: a case study. In Proceedings of the 41st International Conference on Software Engineering: Software Engineering in Practice (ICSE-SEIP '19). IEEE Press, 291–300.
- Nithya Sambasivan Shivani Kapania Hannah Highfill Diana Akrong Praveen Kumar Paritosh Lora Aroyo. 2021. "Everyone wants to do the model work, not the data work": Data Cascades in High-Stakes AI, SIGCHI, ACM (2021)
- Lora Aroyo, Praveen Paritosh. 2021. Uncovering Unknown Unknowns in Machine Learning, https://ai.googleblog.com/2021/02/uncovering-unknown-unknowns-in-machine.html